\documentclass[conference]{IEEEtran}
\IEEEoverridecommandlockouts
\usepackage{cite}
\usepackage{amsmath,amssymb,amsfonts}
\usepackage{algorithmic}
\usepackage{graphicx}
\usepackage{textcomp}
\usepackage{xcolor}
\usepackage{float}
\usepackage{booktabs}
\usepackage{microtype}
\usepackage[T1]{fontenc}
\usepackage{bbding}
\usepackage{pifont}
\usepackage{wasysym}
\usepackage{amssymb}
\usepackage{hyperref}
\def\BibTeX{{\rm B\kern-.05em{\sc i\kern-.025em b}\kern-.08em
    T\kern-.1667em\lower.7ex\hbox{E}\kern-.125emX}}
\begin{document}
{
\title{M-SEVIQ: A Multi-band Stereo Event Visual–Inertial Quadruped-based Dataset for Perception under Rapid Motion and Challenging Illumination\\
}

\author{Jingcheng Cao$^{1,*}$, Chaoran Xiong$^{1,*}$\thanks{$^{*}$Jingcheng Cao and Chaoran Xiong contribute equally to this work.}  , Jianmin Song$^{2}$, Shang Yan$^{2}$\dag,  Jiachen Liu$^{2}$, Ling Pei$^{1}$\thanks{Corresponding author: Ling Pei.} \\ \textsuperscript{1} Shanghai Key Laboratory of Navigation \& Location Based Services, Shanghai Jiao Tong University \\ \textsuperscript{2} Beijing Aerospace Microsystem and Information Technology Research Institute}

}

\maketitle
\begin{abstract}
Agile locomotion in legged robots poses significant challenges for visual perception.  Traditional frame-based cameras often fail in these scenarios for producing blurred images, particularly under low-light conditions.  In contrast, event cameras capture changes in brightness asynchronously, offering low latency, high temporal resolution, and high dynamic range.  These advantages make them suitable for robust perception during rapid motion and under challenging illumination.  However, existing event camera datasets exhibit limitations in stereo configurations and multi-band sensing domains under various illumination conditions.  To address this gap, we present M-SEVIQ, a multi-band stereo event visual and inertial quadruped dataset collected using a Unitree Go2 equipped with stereo event cameras, a frame-based camera, an inertial measurement unit (IMU), and joint encoders.  This dataset contains more than 30 real-world sequences captured across different velocity levels, illumination wavelengths, and lighting conditions.  In addition, comprehensive calibration data, including intrinsic, extrinsic, and temporal alignments, are provided to facilitate accurate sensor fusion and benchmarking.  Our M-SEVIQ can be used to support research in agile robot perception, sensor fusion, semantic segmentation and multi-modal vision in challenging environments.
\end{abstract}


\maketitle
% ----------------- TABLES -----------------
\begin{table*}
\centering
\caption{Comparison of datasets with event cameras.}
\small
\resizebox{\textwidth}{!}
{
\begin{tabular}{@{}lccccccc@{}}
\toprule
\textbf{Dataset}            & \textbf{Event Resolution} & \textbf{Depth Sensor} & \textbf{IMU} & \textbf{Platform}        & \textbf{Environment}   & \textbf{Ground-truth Pose}  & \textbf{NIR light}\\ 
\midrule

D-eDVS              & 128px $\times$ 128px      & RGB-D                 & $\times$       & Handheld                 & Indoor                 & MoCap                       & $\times$\\ 
evbench \cite{2016A}            & 240px $\times$ 180px      & RGB-D                 & $\times$       & Wheeled robot            & Indoor                 & Odometer                    & $\times$\\ 
Mueggler et al. \cite{Mueggler_2017}    & 240px $\times$ 180px      & RGB-D                 & $\times$       & Handheld                 & Indoor/Outdoor         & MoCap                       & $\times$\\ 
MVSEC \cite{Zhu_2018}             & 2 $\times$ 346px $\times$ 260px & LiDAR-16              & $\checkmark$   & Multiple Robots          & Indoor/Outdoor         & MoCap/Cartographer          & $\times$\\ 
UZH-FPV \cite{8793887}           & 346px $\times$ 260px      & $\times$              & $\checkmark$   & Drone                    & Indoor/Outdoor         & MoCap                       & $\times$\\ 
ViViD            & 240px $\times$ 180px      & RGB-D/LiDAR-16        & $\checkmark$   & Handheld                 & Indoor/Outdoor         & MoCap/LeGO-LOAM             & $\times$\\ 
ViViD++ \cite{9760091}          & 240px $\times$ 180px      & RGB-D/LiDAR-64        & $\checkmark$   & Handheld/Car             & Indoor/Outdoor         & MoCap/RTK-GPS               & $\times$\\ 
DSEC \cite{gehrig2021dsecstereoeventcamera}             & 2 $\times$ 640px $\times$ 480px & LiDAR-16              & $\times$       & Car                      & Outdoor                & LeGO-LOAM                   & $\times$\\ 
AGRI-EBV \cite{9561741}         & 240px $\times$ 180px      & RGB-D/LiDAR-16        & $\times$       & Wheeled Robot            & Outdoor                & RTK-GPS                     & $\times$\\ 
TUM-VIE \cite{9636728}          & 2 $\times$ 1280px $\times$ 720px & $\times$              & $\times$       & Handheld/Bike            & Indoor/Outdoor         & MoCap                       & $\times$\\ 
VECtor \cite{9809788}           & 2 $\times$ 640px $\times$ 480px & RGB-D/LiDAR-128      & $\checkmark$   & Handheld/Scooter         & Indoor                 & MoCap/ICP                   & $\times$\\ 
M3ED \cite{10209006}             & 2 $\times$ 1280px $\times$ 720px & LiDAR-64              & $\checkmark$   & Multiple Robots          & Indoor/Outdoor         & Faster-LIO                  & $\times$\\ 
CEAR \cite{Zhu_2024}      & 346px $\times$ 260px      & RGB-D/LiDAR-16        & $\checkmark$   & Agile legged Robot       & Indoor/Outdoor         & MoCap/Faster-LIO            & $\times$\\ \hline
\textbf{M-SEVIQ(ours)} & 2$\times$1280px$\times$720px & RGB-D & $\checkmark$ & Agile legged Robot& Indoor/Outdoor & RTK-GPS & $\checkmark$\\
\bottomrule
\end{tabular}
}
\label{tab:datasets}
\end{table*}
\section{Introduction}\label{MANUSCRIPT}
%\saythanks % added 28-02-2014 Markus Englich

The exceptional mobility of modern robots holds great promise for applications requiring fast navigation through complex environments. Despite considerable progress in robotic locomotion, ~\cite{xiong2025sensingsocialmotionintelligence}, their deployment remains largely restricted to scenarios that do not demand immediate action or high-speed exploration. This constraint impedes their broader adoption in time‑critical tasks such as disaster response, search and rescue, and firefighting. A major challenge lies in degraded perception during rapid movement: motion blur affects RGB images and distortions plague LiDAR point clouds, obstructing accurate pose estimation and scene understanding.

Event cameras—modeled after biological vision—detect logarithmic intensity changes and offer both low latency and high temporal resolution. Moreover, their high dynamic range (HDR) and superior sensitivity to near‑infrared light(NIR) equip robotic systems to handle both low‑light and overexposed scenarios. Although event cameras have gained traction in high‑speed perception domains such as aerial drones and autonomous vehicles, their use in legged robotics remains underexplored~\cite{xiong2025theseanheartratevariationinspired}~\cite{BIG}. When integrated effectively, these sensors can significantly enhance state estimation and terrain perception for legged robots.

However, event cameras also have limitations when used in isolation. Because they rely solely on brightness changes, their output is intrinsically tied to the sensor’s ego‑motion, making it difficult to distinguish between the robot’s movement and changes in the external scene. This ambiguity becomes especially problematic when the robot moves slowly or along textures aligned with its motion direction. Overcoming these challenges—and boosting the performance of agile legged robots in varied environments—calls for algorithms that fuse multiple sensing modalities~\cite{5507190}.

To overcome this challenge, we introduce a novel dataset collected using an advanced sensor suite mounted on the agile quadruped robot, Unitree Go2 . The system consists of stereo event cameras, an RGB‑D camera, an IMU, an RTK module~\cite{8736355}, and twelve joint‑angle sensors. By integrating event and RGB cameras, the system supports perception across both slow and fast movements. Depth information—crucial for precise state estimation and navigation  is captured by the RGB‑D and stereo event cameras. The IMU records high‑frequency angular and acceleration data, unaffected by external dynamic disturbances. Joint angle measurements further enrich odometry via forward kinematics.Our dataset spans a range of velocities and encompasses a diverse array of indoor and outdoor environments under various lighting conditions (e.g., daytime, nighttime, and near-infrared illumination in dark settings). This broad coverage enables a comprehensive evaluation of the resilience and adaptability of perception systems across diverse real-world scenarios. The main contributions of this paper are as follows:  
\begin{enumerate}
  \item We present a diverse event-visual-centric dataset\footnote{Demo and dataset usage: \url{https://anonymous.4open.science/r/202510upinlbs-M-SEVIQ-8FA7}.} encompassing indoor and outdoor scenarios with varying illumination and motion speeds, enabling thorough benchmarking of perception algorithms in real-world environments.
  \item We integrate binocular high-resolution event cameras with proprioceptive sensors and perform both intrinsic and extrinsic calibration, providing a precise sensing framework for dynamic environment perception.
  \item Cross-spectrum event perception is evaluated. Active near-infrared illumination is employed to enhance performance under low-light conditions and improve dataset robustness.
\end{enumerate}

The remainder of this paper is organized as follows. Section 2 reviews related work on event-based vision datasets. Section 3 details the hardware configuration and calibration of our data collection platform. Section 4 describes the various scenarios and sequences included in the proposed dataset. Section 5 presents experimental validation of existing algorithms on the M-SEVIQ dataset, demonstrating its utility. Finally, Section 6 concludes the paper.

\section{Related work}\label{sec:TITLE AND ABSTRACT BLOCK}
This section provides an overview of several key event camera datasets, which are summarized in Table I. The first dataset, introduced in 2014, utilized a 128×128 eDVS event camera ~\cite{5457625} alongside an RGB-D camera. Two years later, a dataset was published ~\cite{2016A} featuring a 240×180 pixel event camera designed for visual navigation of a mobile wheeled robot. In 2017, another dataset ~\cite{Mueggler_2017} was released, offering a comprehensive handheld dataset with motion speed data, serving as a reference for exploring the performance of event cameras (240×180 pixels). These datasets have significantly contributed to the adoption and development of event cameras in research.

Recent advancements have resulted in datasets featuring higher-resolution event cameras. In 2022, the ViViD dataset was upgraded to ViViD++ \cite{9760091}, incorporating a high-resolution event camera (640 × 480 pixels) tailored for driving scenarios. This update included repeated data collection along predefined trajectories at different times, with motion categorized as slow, unstable, or aggressive. The TUM-VIE dataset \cite{9636728} captures stereo event camera recordings (1280 × 720 pixels) across various environments and activities, including walking, running, skating, and biking. Likewise, the VECtor dataset ~\cite{9809788} offers stereo event camera sequences (640 × 480 pixels) from both head-mounted and pole-mounted setups in indoor environments. These high-resolution datasets enhance image clarity, facilitating more detailed and accurate analyses.

Event camera datasets have also been developed for specific research purposes. For example, the DDD17 ~\cite{binas2017ddd17endtoenddavisdriving} and DDD20 \cite{10.1109/ITSC45102.2020.9294515} datasets provide extensive metadata, including vehicle speed, GPS coordinates, and detailed driving dynamics such as steering, throttle, and brake inputs, which enhance the accuracy of steering prediction. The dataset presented in \cite{9201344} addresses place recognition by mounting a DAVIS346 event camera (346 × 260 pixels) on a vehicle, capturing data along the same route at different times. Another dataset, created by ~\cite{9561741}, focuses on agricultural robotics in diverse farming environments using a mobile wheeled robot. Additionally, ~\cite{gehrig2021dsecstereoeventcamera}employed a stereo camera setup, coupled with a 16-channel Velodyne LiDAR and an RTK-GPS system, for automotive driving applications. These specialized datasets enable in-depth exploration of event cameras in targeted areas of research.

A notable trend in event camera datasets is their use in various robotic platforms. For instance, ~\cite{8793887} introduced a dataset focused on the fast-paced dynamics of drone racing. In 2018, ~\cite{Zhu_2018} released the first multi-robot dataset, capturing both indoor and outdoor environments with handheld devices, hexacopters, vehicles, and motorcycles. Additionally, ~\cite{10209006} presented the M3ED dataset, which features event cameras integrated with multiple sensor arrays across various robotic platforms, such as unmanned aerial vehicles, wheeled ground vehicles, and legged robots. While the M3ED dataset includes data from a quadruped robot, Spot, it only captures basic trotting motions and lacks more dynamic movements, such as pronking or bounding gaits, as well as varied environmental conditions.The CEAR dataset includes a wider range of quadruped robot gaits, multi-modal sensor data (including dense depth data), and diverse scenarios collected under different lighting conditions in varied environments. In contrast, our dataset, M-SEVIQ, adopts a more generalized operational state, preserving the diversity of working scenarios and the integration of multi-sensor setups. Building upon this, we selected high-resolution event cameras and incorporated a stereo event visual solution that aligns more closely with the concept of embodied intelligence. Additionally, we actively introduced the near-infrared spectrum to enhance the dataset.

\begin{figure}[ht!]
\begin{center}
		\includegraphics[width=0.8\columnwidth]{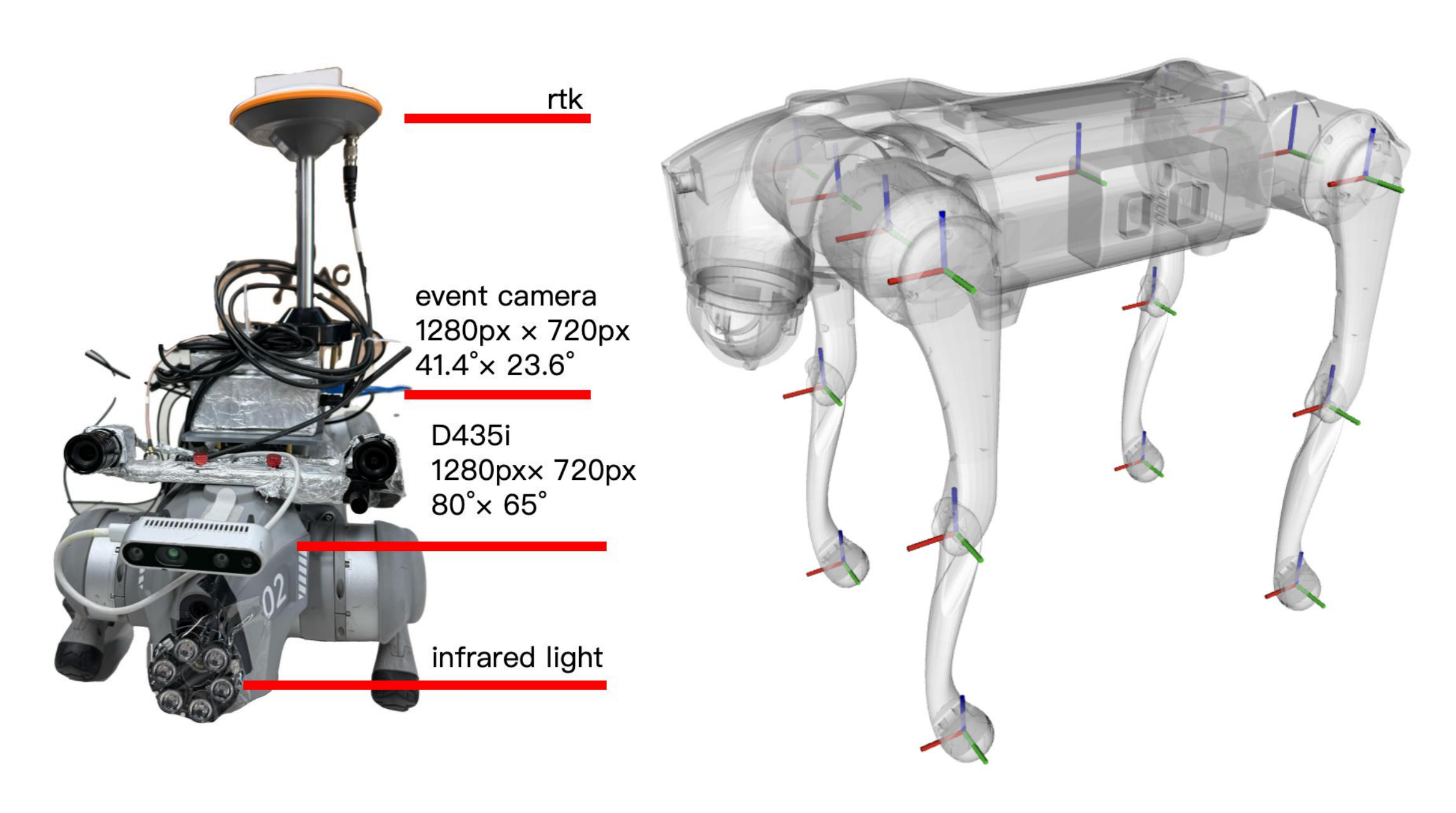}
        \includegraphics[width=0.8\columnwidth]{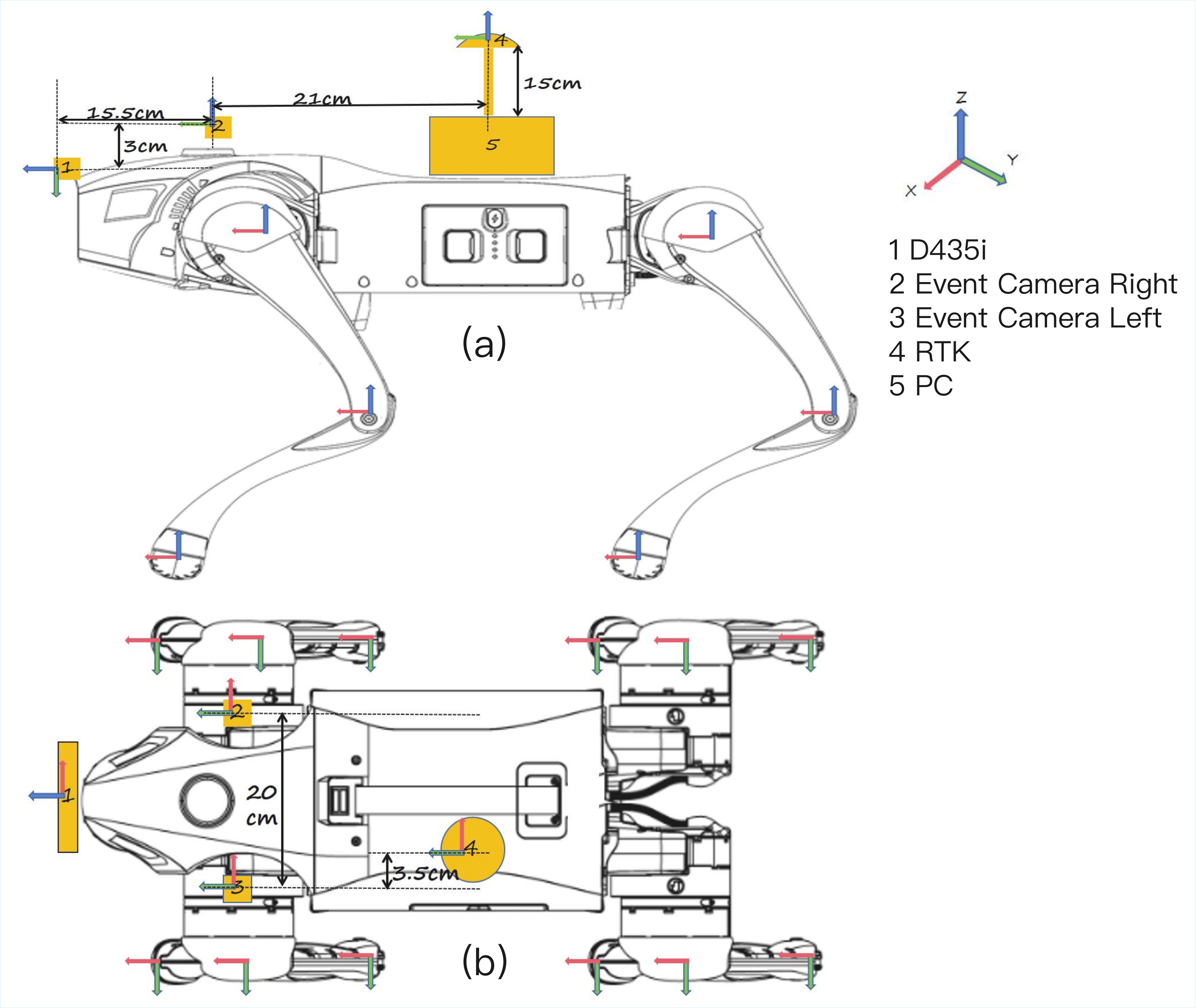}
	\caption{Our Quadruped robot for data collection. (a)Side view of the robot (b)Top view of the robot}
\label{fig:figure_placement}
\end{center}
\end{figure}

\section{Quadruped robot data collection platform}

Fig1 shows the hardware setup, which includes stereo event cameras, a RealSense D435i RGB-D camera with an integrated IMU, and 12 joint encoders on the Unitree Go2 robot. The event camera is selected for its high pixel resolution and excellent near-infrared light sensitivity. All vision sensors and the IMU are securely mounted on the robot using custom 3D-printed fixtures. The Unitree Go2 features 12 actuators that control its limbs, with joint numbering beginning at the front right abduction/adduction joint, followed by the hip and knee flexion/extension joints, and continuing to the left and right front and hind limbs, as illustrated in Fig1. Sensor specifications are detailed in Fig1 too.

\subsection{Sensors Overview}

The Prophesee EVK4 event camera (1280$\times$720 pixels) features a 41.4° horizontal and 23.6° vertical field of view, with a dynamic range of over 120 dB. This camera excels in capturing logarithmic intensity changes, making it highly suitable for rapid perception tasks in dynamic environments. Additionally, the EVK4’s ability to detect near-infrared light enhances its performance in low-light conditions.

The Intel RealSense D435i (1280$\times$720 pixels), with a field of view of 87°×58°, integrates a global shutter RGB sensor and a depth sensor using stereo vision technology. It also includes an integrated 6-axis IMU, enabling accurate motion tracking and improved point-cloud alignment. The D435i captures depth data with sub-2\% accuracy at 2 meters and supports up to 90 frames per second at its highest resolution. The combination of a wide field of view and global shutter sensors makes the D435i ideal for robotic navigation and object recognition in varying lighting conditions.

\subsection{Calibration}

The calibration process encompasses intrinsic calibration of cameras and an IMU, as well as extrinsic calibration across all sensors.  All parameters are available on the dataset website.\\
\textbf{1) Intrinsic:} Intrinsic parameter calibration of event cameras was performed in order to empirically measure focal length, optical center, and distortion coefficients. Standard steps from Prophesee  observe a checkerboard of known dimensions from varying angles. For the RealSense camera, we did not perform additional intrinsic parameter calibration and used the onboard parameters provided by the manufacturer.\\
\textbf{2) Extrinsic:} Extrinsic parameters between the RealSense RGB camera and the robot were determined from the CAD file. For the RealSense RGB and depth cameras, the onboard extrinsic parameters were provided by the manufacturer. The extrinsic calibration process determines the spatial transformation between each pair of sensors. Given the high temporal resolution and low latency of event cameras, the eKalibr toolbox was utilized to identify the extrinsic parameters between the two Prophesee EVK4 event cameras. eKalibr employs a continuous-time optimization framework, leveraging raw event streams to extract grid patterns and estimate extrinsic parameters through bundle adjustment. This approach ensures precise alignment between the event cameras, facilitating accurate spatiotemporal data fusion. The eKalibr significantly reduces calibration time and simplifies the process, while maintaining comparable accuracy. Then the extrinsic parameters for all sensor pairs can be derived using the identified parameters, creating complete spatial relationships within our sensor array.

\begin{figure}[t]
\begin{center}
		\includegraphics[width=0.8\columnwidth]{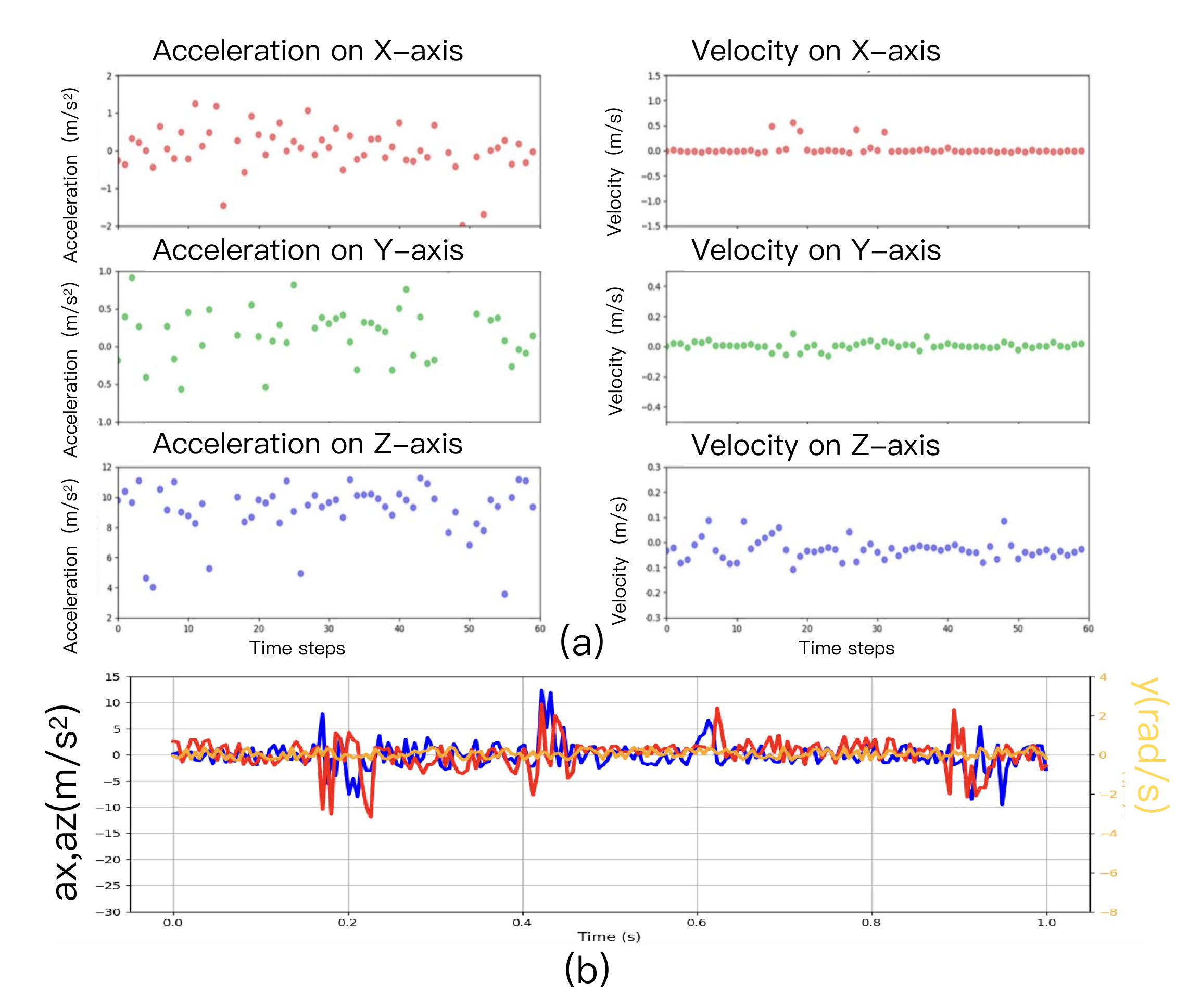}
	\caption{(a) Acceleration and velocity from decoder \\ (b) Acceleration from imu}
\label{fig:figure_placement}
\end{center}
\end{figure}

\subsection{Time Synchronization}

Setting a single timestamp across different sensors is crucial, and hardware synchronization – a single clock source shared by multiple sensors – is often used. However, considering the outdoor data gathering(e.g., motion capture and the robot’s joint encoders), the hardware connection of all sensors was not a practical option for us. On this basis, a specific motion is designed for accurate temporal synchronization through post-processing. 

Before starting the data collection, the robot dog will be held and shake in order to incentivize all IMUs. Then, the robot’s joint encoder data are synchronized with the D435i camera by comparing the angular velocity of the knee joint and the built-in IMU of D435i~\cite{xiong2025a2icalibantinoiseactivemultiimu}. Meanwhile, ROS is used to implement symbolic links (soft connections) for sensor topics to achieve temporal synchronization between D435i and the event camera. During this process, an event camera is used as a reference, and the timestamps of other sensors are offset based on it. 

\begin{figure*}[t]
\begin{center}
		\includegraphics[width=2.0\columnwidth]{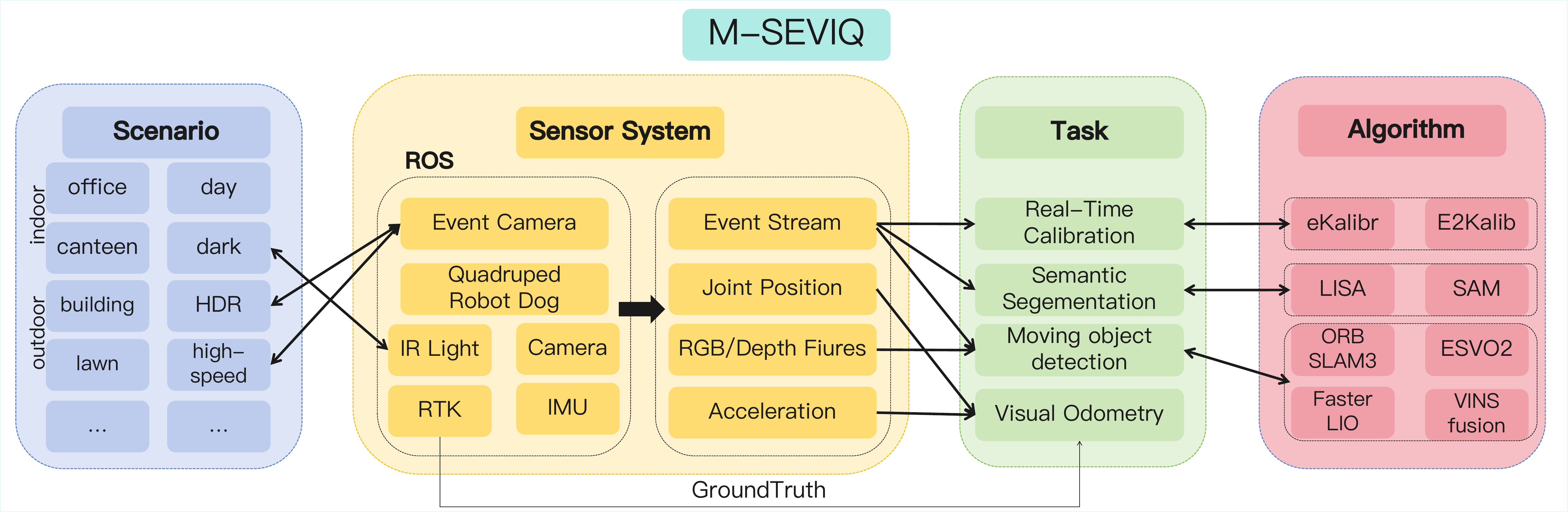}
	\caption{The dataset is systematically organized in a hierarchical manner to facilitate extensive benchmarking. It encompasses a wide range of scenarios, captured using a suite of event-centric sensors. This multi-sensor data provides the foundation for evaluating performance on several perception and navigation tasks. The primary utility of the dataset is demonstrated by benchmarking a set of representative algorithms, establishing a baseline for future research.}
\label{fig:figure_placement}
\end{center}
\end{figure*}

\begin{figure}[t]
\begin{center}
		\includegraphics[width=0.8\columnwidth]{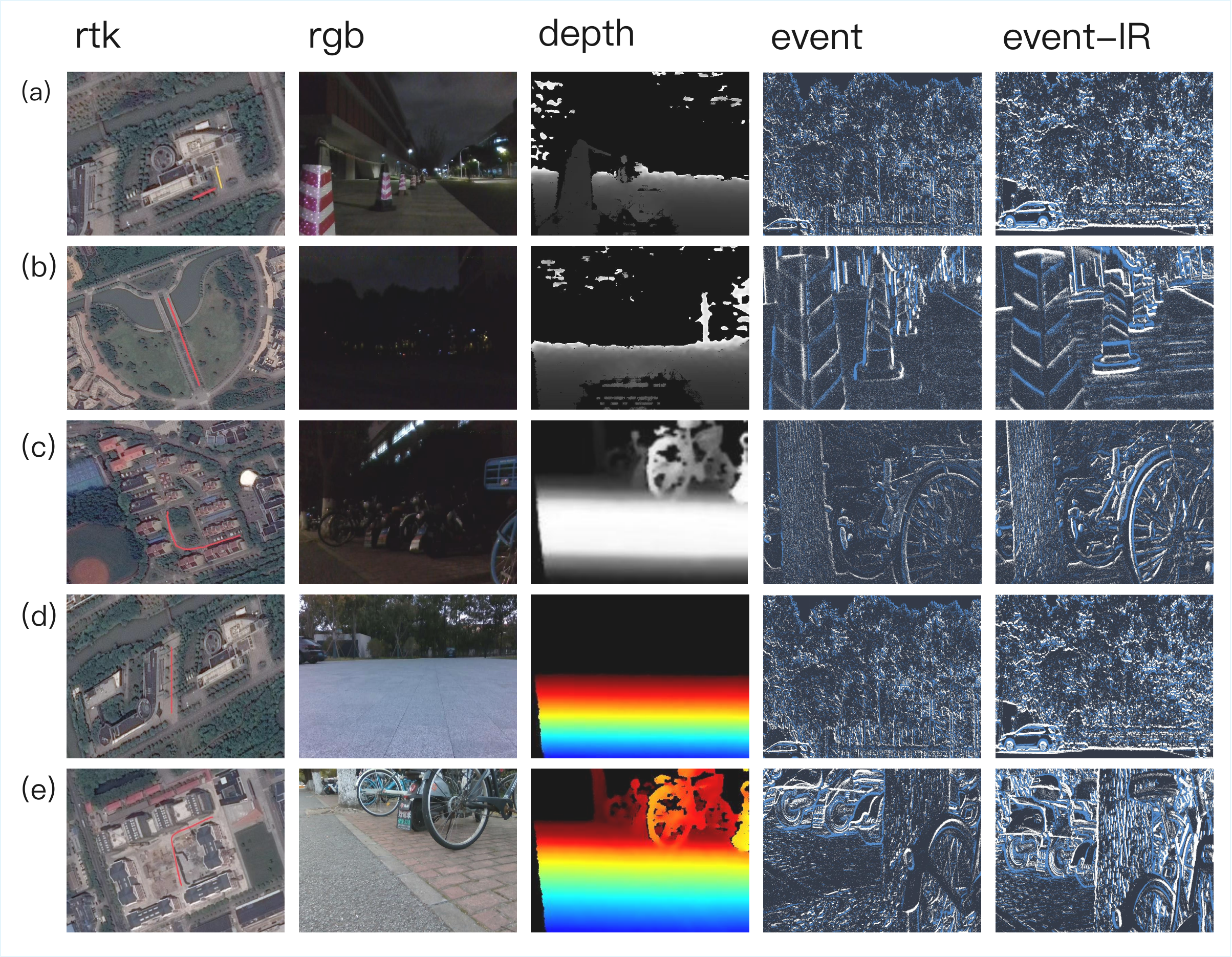}
	\caption{(a) warningpost\_slow\_night; (b) lawn\_slow\_night;(c) bikeparkinglot\_slow\_night; (d) building\_fast\_day; (e) street\_slow\_day}
\label{fig:figure_placement}
\end{center}
\end{figure}

\section{M-SEVIQ}

The M-SEVIQ dataset contains 30+ (will be updated continuously) sequences, including 10 indoor, 20 outdoor. Our dataset aims to evaluate algorithm performance across diverse quadruped velocities and lighting conditions via indoor and outdoor sequences. The name of each sequence reflects the recording environment, lighting conditions, and the velocities of the quadruped robot. For instance, \emph{lawn1\_day\_slow} represents data gathered in a lawn environment during daytime with a slow velocity, while \emph{bikeparking\_blinking\_near-infrared\_comb} indicates data collected in a bike parking lot environment under blinking light conditions and different velocities; meanwhile, near-infrared illumination was actively increased. 

\subsection{Outdoor Sequences}
The outdoor sequences were captured in 10 distinctive outdoor environments, with four sequences per environment.   Each set of four sequences includes variations of two different sets of infrared light under different periods of the day (daytime and nighttime), ensuring a comprehensive representation of real-world settings.     Fig. 4 provides an illustration of all outdoor environments and associated challenges.

\subsection{Indoor Sequences}
The indoor sequences were collected in several diverse environments. In the dining hall, building floor, and office environments, M-SEVIQ recorded two sequences per site, with each sequence showcasing a different set of speeds that can reach 10m/s. Data collection was expanded to encompass a range of lighting conditions: well-lit and dark environments with bright light sources, which is called a high dynamic range (HDR) condition.  We envision that these indoor sequences will enable a comprehensive evaluation of agile quadruped robots across varied indoor environments and under diverse lighting conditions.

In addition to challenges commonly encountered in outdoor environments, such as varying lighting conditions, dynamic objects, and slippery floors, indoor environments present unique obstacles. These include reflections from floors or glass surfaces that can interfere with perception, as well as closely spaced obstacles that may temporarily obstruct the camera's line of sight. Moreover, the presence of repetitive scenes could potentially distract or disrupt the subject's focus.

\subsection{Data Usage and tools}
All the data except RTK were captured by rosbag in Robot Operation System (ROS), and the recorded topics are listed as follows. The RTK data was recorded synchronously with independent time stamp.

\begin{itemize}
\item[$\bullet$]Event Camera:\\
/prophesee/camera\_left/cd\_events\_buffer\\
/prophesee/camera\_right/cd\_events\_buffer\\
\item[$\bullet$]D435i camera:\\
/camera/color/image\_raw \\
/camera/depth/image\_rect\_raw\\
/camera/imu\\
/camera/accel/sample\\
/camera/gyro/sample\\
\item[$\bullet$]Unitree go2 decoder and IMU:\\
/unitree\_go2/low\_state
\end{itemize}
\section{Evaluation: Semantic Segmentation}
The majority of existing event camera datasets typically use established SLAM algorithms~\cite{11037226} ~\cite{7932451}as benchmarks, validating the dataset through tasks such as odometry or obstacle avoidance. However, in this study, in addition to demonstrating the dataset's complexity and high quality, we aim to highlight the extended capabilities of the M-SEVIQ dataset: semantic segmentation, which employs high-resolution event cameras and active near-infrared illumination, moving beyond the scope of traditional event camera datasets. Retrieving accurate semantic information under challenging high dynamic range (HDR) and high-speed conditions remains a significant challenge for image-based algorithms due to severe image degradation. Event cameras have the potential to address these challenges since they feature a significantly higher dynamic range and are resilient to motion blur. To achieve this, we applied existing traditional computer vision algorithms to segment the event camera images.

In the semantic segmentation evaluation, we primarily employed several traditional frame-based segmentation algorithms~\cite{https://doi.org/10.13140/rg.2.2.33195.86560}. While these methods have been widely validated on RGB/RGB-D images, they face performance limitations when applied to event camera outputs, which are high-frequency, sparse, and characterized by low-latency event streams. A noteworthy model in this context is the Segment Anything Model (SAM), which offers a promptable segmentation framework with strong zero-shot transfer capabilities. SAM utilizes an image encoder, a prompt encoder, and a mask decoder, with a modular architecture trained on the large-scale SA-1B dataset, enabling rapid segmentation across a variety of images. Despite SAM’s promising capabilities, there is limited research on directly applying SAM to semantic segmentation of event camera images or event streams. Surprisingly, M-SEVIQ's ability to robustly and universally perform object segmentation has been demonstrated using SAM .

\begin{figure}[t]
\begin{center}
		\includegraphics[width=0.8\columnwidth]{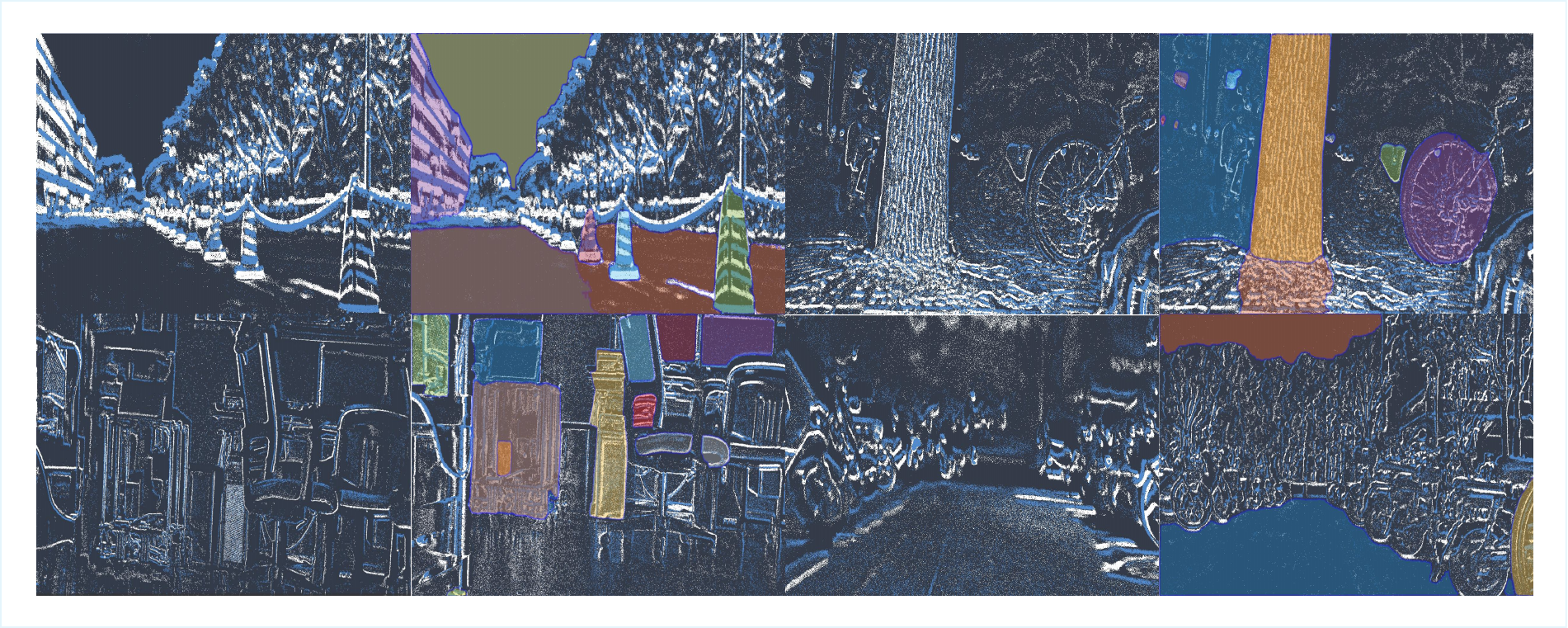}
	\caption{Segment the picture automatically}
\label{fig:figure_placement}
\end{center}
\end{figure}

Although perception systems have made significant advancements in recent years, they still depend on explicit human instruction or predefined categories to identify the target objects before executing visual recognition tasks.       Such systems are unable to actively reason and comprehend implicit user intention. We tested Large Language Instructed Segmentation Assistant (LISA) in our dataset.

In contrast to SAM, LISA is designed explicitly to integrate multimodal large language model (LLM) reasoning capabilities: it accepts both visual inputs and complex language queries, introduces a dedicated token (〈SEG〉) and uses an embedding‑as‑mask paradigm to produce segmentation outputs that reflect implicit user intent, world knowledge and reasoning.      LISA, which inherits the language generation capabilities of multimodal Large Language Models (LLMs) while also possessing the ability to produce segmentation maps, its paradigm of language‑instructed segmentation better aligns with scenarios where semantic queries might involve temporal and modality complexity.

As a potential research direction, we believe that the dataset presented in this paper, which includes complex environments such as rapid motion, nighttime/strong light, and near-infrared conditions, provides a novel and challenging evaluation platform for algorithms in the domain of event camera semantic segmentation.   It also provides a novel and challenging benchmark for both models: SAM as a baseline tool for segmentation in event‑camera imagery, and LISA (or its derivatives) as a promising future pathway for language‑instructed segmentation on event‑camera data.  In future work, we plan to explore fine-tuning SAM or leveraging prompt engineering techniques, utilizing this dataset to evaluate its generalization capabilities on event camera images and to compare its performance with traditional segmentation algorithms, and explore prompt engineering and fine‑tuning pipelines for LISA to evaluate its generalisation and reasoning capabilities in the event‑camera domain.

\begin{figure}[t]
\begin{center}
		\includegraphics[width=0.8\columnwidth]{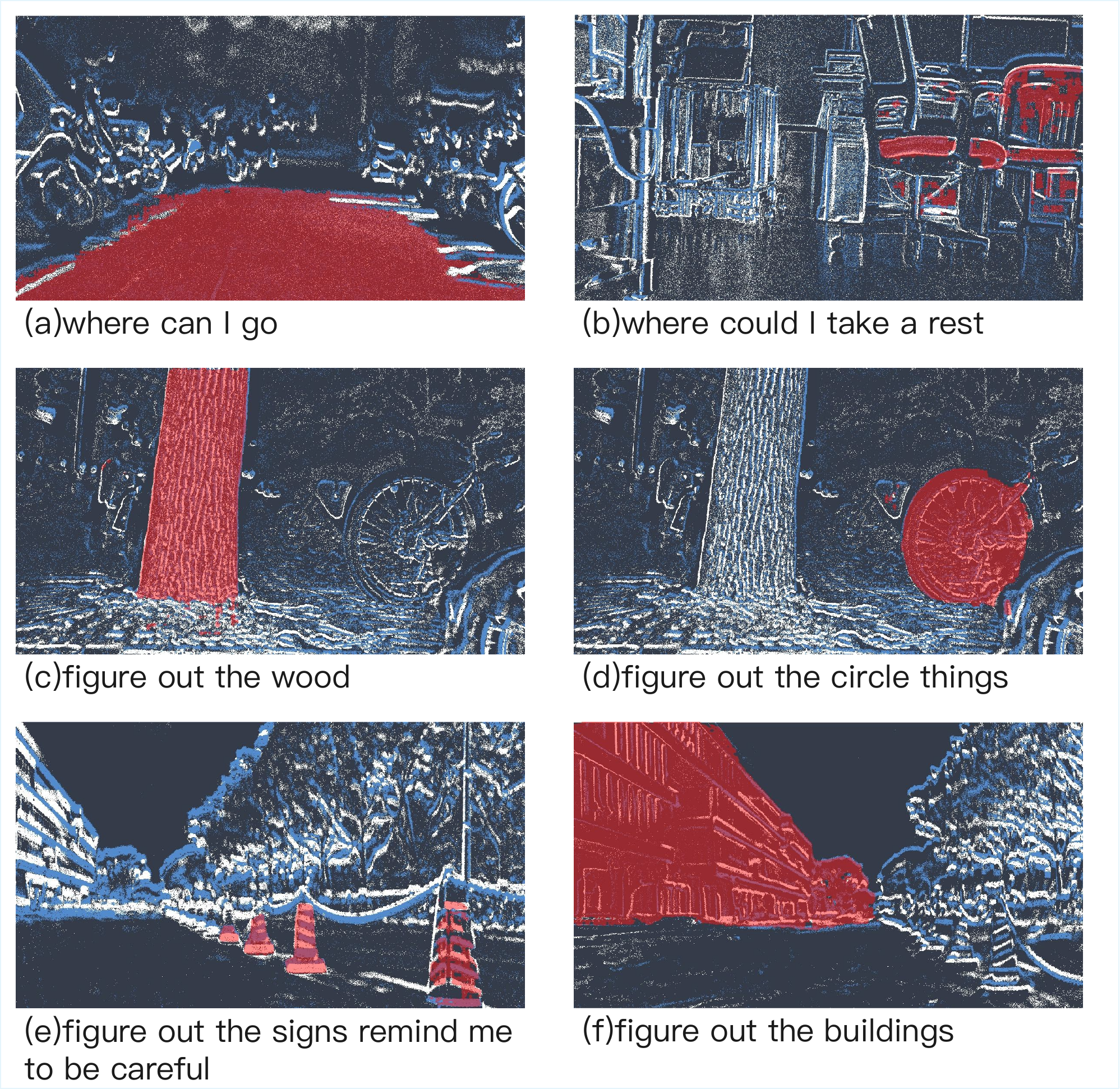}
	\caption{Combined semantic segmentation with LLM by LISA}
\label{fig:figure_placement}
\end{center}
\end{figure}

\section{Conclusions}
In this paper, we introduce the M-SEVIQ, the pioneering dataset designed to investigate the perception system of a dynamic quadruped robot by utilizing event/RGB-D cameras, RTK, IMU, and joint encoders. The M-SEVIQ dataset offers a comprehensive collection of indoor and outdoor sequences under various velocities and lighting conditions, especially the set of stereo event camera and near infrared illumination. We envision the M-SEVIQ dataset as a cornerstone for exploring the rapid perception of event cameras in robotics and multi-sensor fusion research. In future work, we plan to expand the dataset in terms of coverage and number of samples, and develop a quadruped robot platform capable of autonomously performing multi‑sensor fusion calibration tasks at the system level, and develop data-driven methods using M-SEVIQ dataset.

\vspace{12pt}

\end{document}